# Exploring Localization In Bayesian Networks For Large Expert Systems

Yang Xiang**, David Poole† and Michael P. Beddoes*
** Expert Systems Laboratory, Centre for Systems Science
Simon Fraser University, Burnaby, B.C., Canada V5A 1S6, yangx@cs.sfu.ca
† Department of Computer Science, University of British Columbia
Vancouver, B.C., Canada, V6T 1W5, poole@cs.ubc.ca
* Department of Electrical Engineering, University of British Columbia
Vancouver, B.C., Canada, V6T 1W5, mikeb@ee.ubc.ca

## Abstract

Current Bayesian net representations do not consider structure in the domain and include all variables in a *homogeneous* network. At any time, a human reasoner in a large domain may direct his attention to only one of a number of natural subdomains, i.e., there is 'localization' of queries and evidence. In such a case, propagating evidence through a homogeneous network is inefficient since the entire network has to be updated each time. This paper presents multiply sectioned Bayesian networks that enable a (*localization preserving*) representation of natural subdomains by separate Bayesian subnets. The subnets are transformed into a set of permanent junction trees such that evidential reasoning takes place at only one of them at a time. Probabilities obtained are identical to those that would be obtained from the homogeneous network. We discuss attention shift to a different junction tree and propagation of previously acquired evidence. Although the overall system can be large, computational requirements are governed by the size of only one junction tree.

## 1 LOCALIZATION

### 1.1 WHAT IS LOCALIZATION?

We consider the following general context where an expert system is to be used (Figure 1): The human user plays the central role between an expert system and a real world domain. To know the state of the domain (e.g. diagnosis), the user gathers evidence from the domain and enters the evidence along with queries to the expert system. The system may provide a recommendation or may prompt further gathering of information. We would like the system to be efficient in inference computation. How can this aim be realized when the domain is large?

We define informally what we will call *localization* as a property of large domains with respect to human cognitive activity.

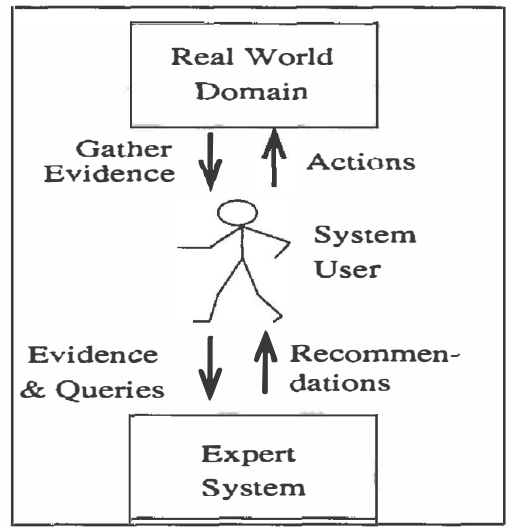

Figure 1: An illustration of the context

> There are fixed 'natural' subdomains. At any time, a human reasoner focuses attention at only one of them. He can acquire evidence from and form queries about only one of them at a time. He may shift attention from one to another from time to time.

Localization can be seen in many large domains. For example, in many medical domains (e.g. neuromuscular diagnosis [Xiang et al. 1991]) medical practitioners acquire information about a patient by history taking, physical examination, and performing a set of specialized tests. Each such activity involves a subdomain which contains possibly dozens of alternative questions, procedures, test sites, etc., and diseases which can be differentiated. Each such subdomain can hold the person's attention for a period of time. During this period, he updates his belief on disease hypotheses based on acquired evidence, carefully weighs the importance of alternative means to gather information under the current situation, and selects the best alternative to perform next. Although one subdomain may have an influence on another, the influence is summarized by the common disease hypotheses which they both can (partially) differentiate.

We distinguish 'interesting' variables from 'relevant' variables. Call the set $U$ of variables in a subdomain 'interesting' to the human reasoner if this subdomain captures his current attention. A set $V$ of variables outside the current subdomain may have a bearing on $U$ due to two reasons. First, the background knowledge on $U$ may provide partial background knowledge on $V$. Second, obtained evidence on $U$ may change one's belief on $V$. Thus $V$ is 'relevant' to $U$ but is not currently 'interesting' to the reasoner. However, we can often find a set $I$ ($I \subset U$) of variables which summarizes all the influence on $U$ from $V$ such that $V$ can be discarded from the reasoner's current attention. This issue is treated more formally in Section 4.1.

### 1.2 IS LOCALIZATION USEFUL?

When localization exists, a large domain can be represented in an expert system according to the natural subdomains. Each subdomain is represented by a subsystem. During a consultation session, only the subsystem corresponding to the current subdomain consumes computational resources. This subsystem is called *active*. When a user's attention shifts to a different subdomain, the evidence acquired in the previously active subsystem can be absorbed by the newly active subsystem through summarizing variables. Since no computational resources are consumed by the inactive subsystems, computational savings can be obtained without loss of inference accuracy.

Our observation of localization was made based on our experience in the domain of neuromuscular diagnosis [ib.]. We believe that it is common in many large domains. How to exploit it when it arises, how to construct the above ideal representation in the context of Bayesian networks, and how to guarantee the correctness of inference in the representation is the subject of this paper.

## 2 EXPLORE LOCALIZATION IN BAYESIAN NETS

### 2.1 BACKGROUND

Cooper [1990] has shown that probabilistic inference in a general Bayesian net is NP-hard. Several different approaches have been pursued to avoid combinatorial explosion for typical cases, and thus to reduce computational cost. Two classes of approaches can be identified. One class explores *approximation* [Henrion 1988; Pearl 1988; Jensen and Andersen 1990]. Another class explores specificity in computing *exact* probabilities [Pearl 1986; Heckerman 1990a; Lauritzen and Spiegelhalter 1988; Jensen, Lauritzen and Olesen 1990a; Baker and Boult 1990; Suermondt, Cooper and Heckerman 1990].

This paper takes the exact approach. For general but sparse nets, efficient computation has been achieved by creating a secondary directed [Lauritzen and Spiegelhalter 1988], or undirected clique tree (junction tree) [Jensen, Lauritzen and Olesen 1990a] structure, which also offers the advantage of trading compile time with running time for expert systems. Both methods and many others are based on a net representation which does not consider domain structure and lumps all variables into a *homogeneous* network.

Pruning Bayesian nets with respect to each query instance is another exact method with savings in computational cost [Baker and Boult 1990]. The method does not support incremental evidence (i.e. all evidence must be entered at one time).

Heckerman [1990b] partitions Bayesian nets into small groups of naturally related variables to ease the construction of large networks. But once the construction is finished, the run time representation is still homogeneous.

Suermondt, Cooper and Heckerman [1990] combine cutset conditioning with the clique tree method and convert the original net into a set of clique trees to obtain computational savings. The cutset is chosen mainly based on net topology. It does not lead to the exploration of localization in general.

### 2.2 'OBVIOUS' WAYS TO EXPLORE LOCALIZATION

#### Splitting homogeneous nets

One obvious way to explore localization is to split a homogeneous Bayesian net into a set of subnets according to localization. Each subnet can then be used as a separate computational object. This is not always workable as is shown by the following example.

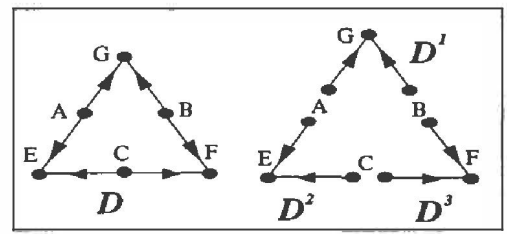

Figure 2: Left: a DAG $D$. Right: A set of subnets formed by sectioning $D$.

Suppose the directed acyclic graph (DAG) $D$ in Figure 2 is split according to localization into $\{D^1, D^2, D^3\}$. Suppose variable $G$ is instantiated by evidence. According to d-separation [Pearl 1988], now both paths between $E$ and $F$ are active. Therefore, in order to pass a new piece of evidence on $E$ to $F$, the joint distribution on $\{B, C\}$ needs to be passed from $\{D^1, D^2\}$ to $D^3$ [1]. However, this is not possible because neither $D^1$ nor $D^2$ contains this joint distribution. This shows that arbitrarily partitioning Bayesian nets causes loss of information and is incorrect in general.

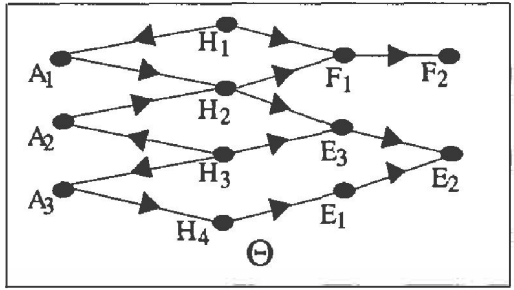

Figure 3: An unsectioned Bayesian net.

### Splitting the junction tree

Another obvious way to explore localization is to preserve localization within subtrees of a junction tree [Jensen, Lauritzen and Olesen 1990a] by clever choice in triangulation and junction tree construction. If this can be done, the junction tree can be split and each subtree can be used as a separate computational object. The following example shows that this is also not always workable. Consider the DAG $\Theta$ in Figure 3. Suppose variables in the DAG form three naturally related groups which satisfy localization:

$$G_1 = \{A_1, A_2, A_3, H_1, H_2, H_3, H_4\}$$
$$G_2 = \{F_1, F_2, H_1, H_2\}$$
$$G_3 = \{E_1, E_2, E_3, H_2, H_3, H_4\}$$

We would like to construct a junction tree which would preserve the localization within three subtrees. The graph $\Phi$ in Figure 4 is the moral graph of $\Theta$. Only the cycle $A_3 - H_3 - E_3 - E_1 - H_4 - A_3$ needs to be triangulated. There are six distinct ways of triangulation out of which only two do not mix nodes in different groups. The two triangulations have the link $(H_3, H_4)$ in common but they do not make a significant difference in the following analysis. The graph $\Lambda$ in Figure 4 shows one of the two triangulations. All the cliques in $\Lambda$ appear as nodes of graph $\Gamma$.

The junction tree $\Gamma$ does not preserve localization since cliques 3, 4, 5 and 8 correspond to group $G_1$ but are connected via cliques 6 and 7 which contain $E_3$ from group $G_3$. This is unavoidable. When there is evidence for $A_1$ or $A_2$ in $\Lambda$, updating the belief in group $G_3$ requires passing the joint distribution of $H_2$ and $H_3$. But updating the belief in $A_3$ only requires passing the marginal distribution of $H_3$. That is to say, updating the belief in $A_3$ needs less information than group $G_3$. In the junction tree representation, this becomes a path from cliques 3, 4 and 5 to clique 8 via cliques 6 and 7.

[1] Passing only the marginal distributions on $B$ and on $C$ is *not* correct.

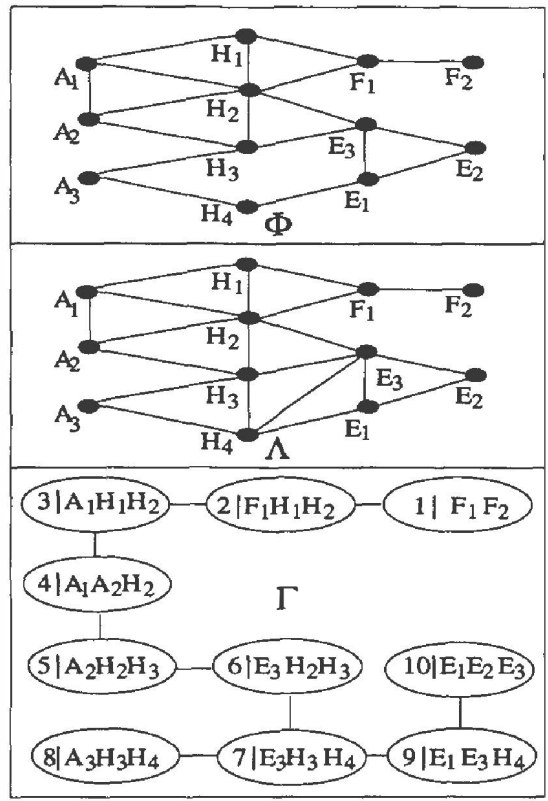

Figure 4: $\Phi$: the moral graph of $\Theta$ in Figure 3. $\Lambda$: a triangulated graph of $\Phi$. $\Gamma$: the junction tree constructed from $\Lambda$.

In general, let $X$ and $Y$ be two sets of variables in the same natural group, and let $Z$ be a set of variables in a distinct group. Suppose the information exchange between pairs of them requires the exchange of distribution on sets $I_{XY}$, $I_{XZ}$ and $I_{YZ}$ of variables respectively. Sometime $I_{XY}$ is a subset of both $I_{XZ}$ and $I_{YZ}$. When this is the case, a junction tree representation will always indirectly connect cliques corresponding to $X$ and $Y$ through cliques corresponding to $Z$ if the method in Jensen, Lauritzen and Olesen [1990a] is followed.

### A brute force method

There is, however, a way around the problem with a brute force method. In the above example, when there is evidence for $A_1$ or $A_2$, the brute force method pretends that updating the belief in $A_3$ needs as much information as $G_3$. What one does is to add a dummy link $(H_2, A_3)$ to the moral graph $\Phi$ in Figure 4. Then triangulating the augmented graph gives the graph $\Lambda'$ in Figure 5. The resultant junction tree $\Gamma'$ in Figure 5 does have three subtrees which correspond to the three groups desired. However, the largest cliques now have size four instead of three as before. In the binary case, the size of the total state space is 84 instead of 76 as before.

In general, the brute force method preserves natural localization by congregation of a set of interfacing nodes (nodes $H_2, H_3, H_4$ above) between natural groups. In this way, the joint distribution on interfacing nodes can be passed between groups, and preservation of localization and preservation of tree structure can be compatible. However, in a large application domain with the original network sparse, this will greatly increase the amount of computation in each group due to the exponential enlargement of the clique state space. The required increase of computation could outweigh the savings gained by exploring localization in general.

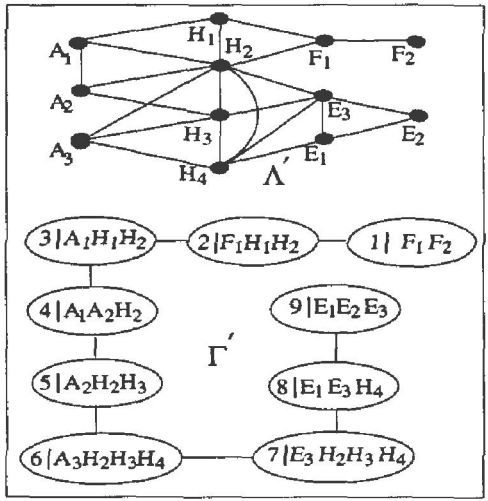

Figure 5: $\Lambda'$ is a triangulated graph. $\Gamma'$ is a junction tree of $\Lambda'$.

The trouble illustrated in the above two situations can be traced to the tree structure of a junction tree representation which requires a single path between any two cliques in the tree. In the normal triangulation case, one has small cliques but one loses localization. In the brute force case, one preserves localization but one does not have small cliques. To summarize, the preservation of natural localization and small cliques can *not* coexist by the method of Andersen et al. [1989] and Jensen, Lauritzen and Olesen [1990a]. It is claimed here that this is due to a single information channel between local groups of variables. This paper present a representation which, by introducing multiple information channels between groups and by exploring conditional independence, allows passing the joint distribution on a set of interfacing variables between groups by passing only marginal distributions on subsets of the set.

## 3 MULTIPLY SECTIONED BAYESIAN NETS

This section introduces a knowledge representation formalism, *Multiply Sectioned Bayesian Networks* (MSBNs), as our solution to explore localization.

We want to partition a large domain according to natural localization into subdomains such that each can be represented separately by a Bayesian subnet. Each subnet then stands as a computational object, and different subnets cooperate with each other during attention shift by exchanging a small amount of information between them. We call such a set of subnets a MSBN. The construction of a MSBN can be formulated conceptually in the opposite direction. Suppose the domain has been represented with a homogeneous network. We frequently refer to a homogeneous net as an *UnSectioned Bayesian network* (USBN). A MSBN is a set of Bayesian subnets resulted from the sectioning of the corresponding USBN.

For example, the DAG $\Theta$ in Figure 3 is sectioned into $\{\Theta^1, \Theta^2, \Theta^3\}$ in Figure 6 according to the localization described in Section 2.2. A variable shared by 'adjacent' subnets appears in both subnets. The set of shared variables is subject to a technical constraint, in addition to localization, as will be discussed in Section 4.

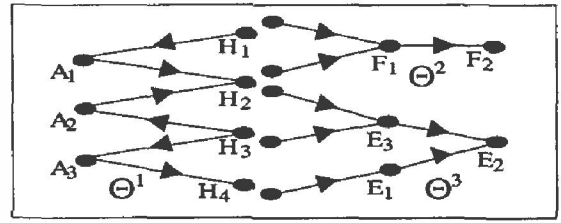

Figure 6: The set $\{\Theta^1, \Theta^2, \Theta^3\}$ forms a MSBN for the USBN $\Theta$ in Figure 3.

### 3.1 TRANSFORMATION OF MSBNS INTO JUNCTION FORESTS

The junction tree representation [Andersen et al. 1989; Jensen, Lauritzen and Olesen 1990a] allows efficient computation for general but sparse networks in expert systems. Thus it is desirable to transform each subnet of a MSBN into a junction tree. The resultant set of junction trees is called a *junction forest*.

For example, the MSBN $\{\Theta^1, \Theta^2, \Theta^3\}$ in Figure 6 is transformed into the junction forest $\{\Gamma^1, \Gamma^2, \Gamma^3\}$ in Figure 7. Omit the ribbed bands for the moment which will be introduced shortly.

In order to propagate evidence between junction trees during attention shift, information channels need to be created between them. As discussed in Section 2.2, multiple channels are required in general to preserve both localization and a small clique size. These channels are called *linkages* between the two junction trees being connected. Each linkage connects two cliques in different trees. The two cliques being connected are called the *host cliques* of the linkage. A linkage is the intersection of its two host cliques. Host cliques are selected such that the union of linkages is the set of interfacing variables between the corresponding subnets, and each linkage is maximal. With multiple linkages created between junction trees, we have a *linked junction forest*.

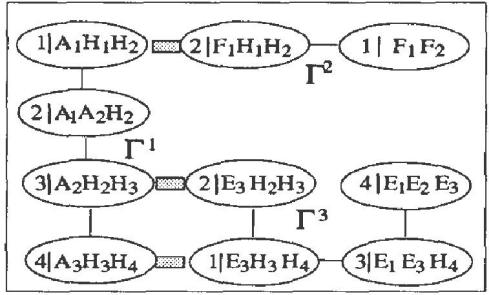

Figure 7: The set $\{\Gamma^1, \Gamma^2, \Gamma^3\}$ is a junction forest transformed from the MSBN $\{\Theta^1, \Theta^2, \Theta^3\}$ in Figure 6. Linkages between junction trees are shown by ribbed bands.

For example, in Figure 7, linkages between junction trees are indicated with ribbed bands connecting the corresponding host cliques. There is only one linkage between clique 1 of $\Gamma^1$ and clique 2 of $\Gamma^2$, namely, $\{H_1, H_2\}$. The two linkages between $\Gamma^1$ and $\Gamma^3$ are $\{H_2, H_3\}$ and $\{H_3, H_4\}$.

Up to here, we have only discussed the manipulations of graphical structures of a MSBN. As other approaches based on secondary structures [Lauritzen and Spiegelhalter 1988; Jensen, Lauritzen and Olesen 1990a], there needs to be a corresponding conversion from the probability distribution of the MSBN to the belief table (potential) of the linked junction forest. Readers are referred to Xiang, Poole and Beddoes [1992] for details regarding the conversion.

### 3.2 EVIDENTIAL REASONING

After a linked junction forest is created, it becomes the permanent representation of the corresponding MSBN. The evidential reasoning during a consultation session will be performed solely in the junction forest.

Due to localization, only one junction tree in a junction forest is active during evidential reasoning. When new evidence becomes available to the currently active junction tree, it is entered and the tree is made consistent. The operations to enter evidence and to maintain consistency within a junction tree are the same as Jensen, Lauritzen and Olesen [1990a]. We only maintain consistency in the currently active tree. All the probabilities in this tree are the same as in a globally consistent junction forest [Xiang, Poole and Beddoes 1992]. It is this feature of MSBNs/junction forests that allows the exploitation of localization.

When the user shifts attention from the currently active tree to a 'destination' tree, all previously acquired evidence is absorbed through an operation **ShiftAttention**. The operation swaps in and out sequentially a chain of 'intermediate' junction trees between the currently active tree and the destination tree. It has been shown [ib.] that, with a properly structured junction forest, the following is true.

> Start with any active junction tree in a globally consistent junction forest. Repeat the following cycle a finite number of times:
> 1. Enter evidence to the currently active tree and make the tree consistent a finite number of times.
> 2. Use **ShiftAttention** to shift attention to any destination tree.
> 
> The marginal distributions obtained in the final active tree are identical to those of a globally consistent forest.

The above property shows the most important characterization of MSBNs and junction forests, namely, the capability of exploiting localization to reduce the computational cost. Note that the above statement only requires the initial global consistency of the junction forest.

With localization, the user's interest and new evidence remain in the sphere of one junction tree for a period of time. Thus the time and space requirement, while reasoning within a junction tree, is bounded above by what is required by the largest junction tree. The judgments obtained take into account all the *relevant* background knowledge and evidence. Compared to the USBN and the single junction tree representation where each piece of evidence has to be propagated through the entire system, this leads to computational savings.

When the user shifts interest to another set of variables contained in a different destination tree, only the intermediate trees need to be updated. The time required is linear to the number of intermediate trees and to the number of linkages between each pair of neighbours [ib.]. No matter how large the entire junction forest, the time requirement for attention shift is fixed once the destination tree and mediating trees are fixed. The space requirement is upper bounded by what is needed by the largest junction tree. With localization, the computational cost for attention shift is incurred only occasionally.

Given the above analysis, the computational complexity of evidential reasoning in a MSBN with $\beta$ subnets of equal size is about $1/\beta$ of the corresponding USBN system given localization. The actual time require-ment is a little more than $1/\beta$ due to the computation required for attention shift. The actual space requirement is a little more than $1/\beta$ due to the repetition of interfacing nodes.

## 4 TECHNICAL ISSUES

Section 2.2 has shown that we cannot divide a homogeneous Bayesian net or its junction tree arbitrarily in order to explore localization. This section discusses major technical issues in the MSBN/junction forest representation.

### 4.1 INTERFACE BETWEEN SUBNETS

Localization does not dictate exactly what should be the boundary between different subnets. The intuitive criterion is that the interface should allow evidence acquired to be propagated to adjacent subnets during attention shift by a small amount of information exchange. We define d-sepset as the criterion of interface, which makes use of Pearl's d-separation concept [Pearl 1988]. We denote the union $D$ of DAGs $D^1$ and $D^2$ by $D = D^1 \sqcup D^2 = (N^1 \cup N^2, E^1 \cup E^2)$.

**Definition 4.1 (d-sepset)** *Let $D = D^1 \sqcup D^2$ be a DAG. The set of nodes $I = N^1 \cap N^2$ is a **d-sepset** between subDAG $D^1$ and $D^2$ if the following condition holds.*

*For every $A_i \in I$ with its parents $\pi_i$ in $D$, either $\pi_i \subseteq N^1$, or $\pi_i \subseteq N^2$.*

*Elements of a d-sepset are called **d-sepnodes**. When the above condition holds, $D$ is said to be **sectioned** into $\{D^1, D^2\}$.*

The following theorem and corollary [Xiang, Poole and Beddoes 1992][2] say that a d-sepset d-separates subnets in a MSBN and is a sufficient information channel.

**Theorem 4.2** *Let a DAG $D$ be sectioned into $\{D^1, D^2\}$ and $I = N^1 \cap N^2$ be a d-sepset. $I$ d-separates $N^1 \setminus I$ from $N^2 \setminus I$.*

**Corollary 4.3** *Let $(D, P)$ be a Bayesian net, $D$ be sectioned into $\{D^1, D^2\}$, and $I = N^1 \cap N^2$ be the d-sepset. When evidence is available at variables in $N^1$, the propagation of the joint distribution on $I$ from $D^1$ to $D^2$ is sufficient in order to obtain posterior distribution on $N^2$.*

### 4.2 OVERALL STRUCTURE OF MSBNS

The d-sepset criterion concerns with the interface between each pair of subnets. This is not sufficient for a workable MSBN.

---

[2]They are simplified here to the MSBN of two subnets.

For example, the sectioning in Figure 2 satisfies the d-sepset condition, but the resultant MSBN does not guarantee correct inference. This is because the sectioning has an unsound overall organization of subnets. Intuitively, the overall structure of a MSBN should ensure that evidence acquired in any subnet be able to propagate to a different subnet by a unique chain of subnets. In the example of Figure 2, after a piece of evidence is available on $G$, a new piece of evidence on $E$ has to propagate to $F$ through two different chains of subnets $D^2 - D^3$ and $D^2 - D^1 - D^3$. This violates the above requirement and causes the problem. The issue of overall structure is treated formally in [ib.].

### 4.3 MORALI-TRIANGULATION BY LOCAL COMPUTATION

Transformation of a MSBN into a junction forest requires moralization and triangulation conceptually the same way as the other approaches based on secondary structures [Lauritzen and Spiegelhalter 1988; Andersen et al. 1989; Jensen, Lauritzen and Olesen 1990a]. However, in the MSBN context, the transformation can be performed globally or by local computation at the level of the subnets. The global computation performs moralization and triangulation in the same way as the other approaches with care not to mix the nodes of distinct subnets into one clique. An additional mapping of the resultant moralized and triangulated graph into subgraphs corresponding to the subnets is needed. But where space saving is concerned, local computation is desired.

Since the number of parents for a d-sepnode may be different for different subnets, the moralization in MSBN cannot be achieved by 'pure' local computation in each subnet. Communication between the subnets is required to ensure that the parents of d-sepnodes are moralized identically in different subnets.

The criterion of triangulation in a MSBN is to ensure the 'intactness' of a resulting hypergraph from the corresponding homogeneous net. Problems arise if one insists on triangulation by local computation at the level of subnets. One problem is that an inter-subnet cycle will be triangulated in the homogeneous net, but the cycle cannot be identified by examining each of the subnets involved individually. Another problem is that cycles involving d-sepnodes may be triangulated differently in different subnets. The solution is to let the subnets communicate during triangulation. Since moralization and triangulation both involve adding links and both require communication between subnets, the corresponding local operations in each subnet can be performed together and messages to other subnets can be sent together. Therefore, operationally, moralization and triangulation in MSBN are not separate steps as in the single junction tree representation. The corresponding integrated operation is termed *morali-triangulation* to reflect this fact.

For example, the MSBN in Figure 6 is morali-

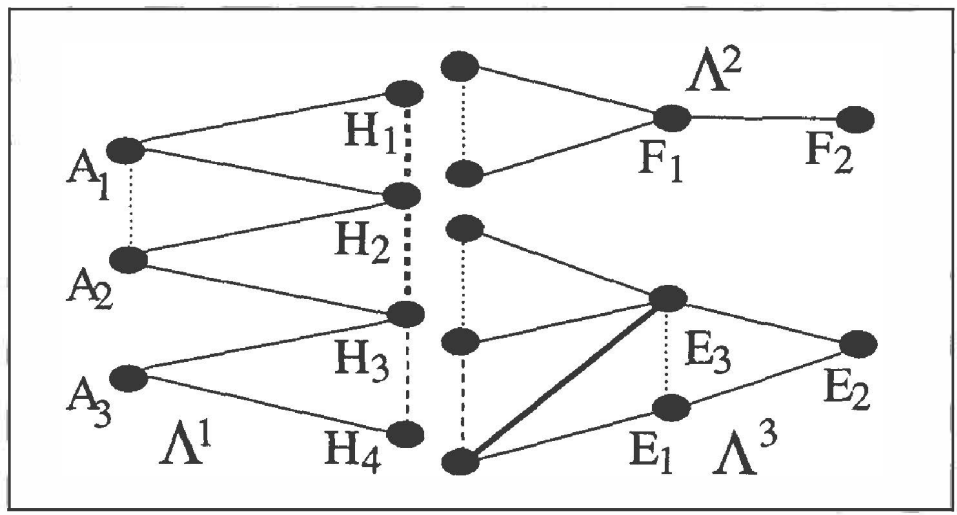

Figure 8: Morali-triangulated graphs of the MSBN in Figure 6. The meaning of the different line-types is explained in Section 4.3.

triangulated to the graphs in Figure 8. Thin solid lines (e.g. $(A_1, H_1)$) are from the original arcs by dropping directions. Thin dotted lines (e.g. $(A_1, A_2)$) are links added by local moralization. Thick dotted lines (e.g. $(H_1, H_2)$) are 'moral' links added through communication. Thin dashed lines (e.g. $(H_3, H_4)$) are added through communication for triangulation. The thick solid line $((E_3, H_4))$ is added by local triangulation. A formal treatment and an algorithm for morali-triangulation are given in Xiang, Poole and Beddoes [1992].

### 4.4 PROPAGATING INFORMATION THROUGH MULTIPLE LINKAGES

Propagating information between junction trees of a junction forest is required in two different situations: belief initialization and evidential reasoning. In both cases, information needs to be propagated between junction trees of a junction forest through multiple linkages. Care is to be taken against potential errors.

Belief initialization serves the same purpose as in other approaches based on secondary structures [Lauritzen and Spiegelhalter 1988; Jensen, Lauritzen and Olesen 1990a]. It establishes the global consistency before any evidence is available. This requires the propagation of knowledge stored in each junction tree to the rest of the forest. When doing so, redundant information could be passed through multiple linkages. We must make sure that the information is passed only once.

In evidential reasoning, evidence acquired in one junction tree needs to be propagated to the destination tree during attention shift. The potential error in this case takes a different form from the case of initialization. Passing information through multiple linkages from one junction tree to another can 'confuse' the receiving tree such that the correct consistency between the two cannot be established.

Detailed illustrations of these potential problems and the operations which avoid them are given in Xiang, Poole and Beddoes [1992].

## 5 CONCLUSION

This paper overviews MSBNs and junction forests as a flexible knowledge representation and as an efficient inference formalism. This formalism is suitable for expert systems which reason about uncertain knowledge in large domains where localization exists.

MSBNs allow partitioning of a large domain into smaller natural subdomains such that each of them can be represented as a Bayesian subnet, and can be tested and refined individually. This makes the representation of a complex domain easier for knowledge engineers and may make the resultant system more natural and more understandable to system users. The modularity facilitates implementation of large systems in an incremental fashion. When partitioning, a knowledge engineer has to take into account the technical constraints imposed by MSBNs which are not very restrictive.

Each subnet in the MSBN is transformed into a junction tree such that the MSBN is transformed into a junction forest where evidential reasoning takes place. Each subnet/junction tree in the MSBN/junction forest stands as a separate computational object. Since the representation allows transformation by local computation at the level of subnets, and allows reasoning to be conducted with junction trees, the space requirement is governed by the size of the largest subnet/junction tree. Hence large applications can be built and run on relatively small computers wherever hardware resources are of concern. This was, in fact, our original motivation for developing the MSBN representation.

During a consultation session, the MSBN representation allows only the 'interesting' junction tree to be loaded while the rest of the forest remains inactive and uses no computational resources. The judgments made on variables in the active tree are consistent with all the knowledge available, including both prior knowledge and all the evidence contained in the entire forest. When the user's attention shifts, inactive trees can be made active and previous accumulation of evidence is preserved. This is achieved by passing the joint beliefs on d-sepsets. The overall computational resources required are governed by the size of the largest subnet, and not by the size of the application domain.

The MSBN has been applied to an expert system PAINULIM for diagnosis of neuromuscular diseases characterized by a painful or impaired upper limb [Xiang et al. 1991].

The MSBN representation makes the localization assumption about the large domain being represented. Our justification of the generality of localization has been intuitive and has been based on our experience in PAINULIM. We are prepared to test its generality in other large domains.

### Acknowledgements

This work is supported by Operating Grants A3290, OGPOO44121 and OGP0090307 from NSERC, and CRD3474 from the Centre for Systems Science at SFU. We are grateful to Stefan Joseph and anonymous reviewers for helpful comments to an earlier draft.